\documentclass[10pt,twocolumn,letterpaper]{article}

\usepackage{cvpr}              
\usepackage{multirow}
\usepackage{graphicx}
\usepackage{amsmath}
\usepackage{amssymb}
\usepackage{booktabs}
\usepackage[accsupp]{axessibility}

\usepackage[pagebackref,breaklinks,colorlinks]{hyperref}

\usepackage[capitalize]{cleveref}
\crefname{section}{Sec.}{Secs.}
\Crefname{section}{Section}{Sections}
\Crefname{table}{Table}{Tables}
\crefname{table}{Tab.}{Tabs.}

\def\cvprPaperID{7108} 
\def\confName{CVPR}
\def\confYear{2023}

\begin{document}

\title{Hierarchical B-frame Video Coding Using Two-Layer CANF\\without Motion Coding}

\author{David Alexandre$^{1}$, Hsueh-Ming Hang$^{2}$, Wen-Hsiao Peng$^{3}$\\
$^{1}$Electrical Engineering and Computer Science International Graduate Program\\
$^{2}$Institute of Electronics\\
$^{3}$Department of Computer Science\\
National Yang Ming Chiao Tung University\\
Hsinchu, Taiwan\\
{\tt\small \{davidalexandre.eed05g,hmhang\}@nctu.edu.tw, wpeng@g2.nctu.edu.tw}
}
\maketitle

\begin{abstract}
   Typical video compression systems consist of two main modules: motion coding and residual coding. This general architecture is adopted by classical coding schemes (such as international standards H.265 and H.266) and deep learning-based coding schemes. We propose a novel B-frame coding architecture based on two-layer Conditional Augmented Normalization Flows (CANF). It has the striking feature of not transmitting any motion information. 
   Our proposed idea of video compression without motion coding offers a new direction for learned video coding. Our base layer is a low-resolution image compressor that replaces the full-resolution motion compressor. The low-resolution coded image is merged with the warped high-resolution images to generate a high-quality image as a conditioning signal for the enhancement-layer image coding in full resolution. One advantage of this architecture is significantly reduced computational complexity due to eliminating the motion information compressor. In addition, we adopt a skip-mode coding technique to reduce the transmitted latent samples. The rate-distortion performance of our scheme is slightly lower than that of the state-of-the-art learned B-frame coding scheme, B-CANF, but outperforms other learned B-frame coding schemes. However, compared to B-CANF, our scheme saves 45\% of multiply–accumulate operations (MACs) for encoding and 27\% of MACs for decoding. The code is available at https://nycu-clab.github.io.  
\end{abstract}


\section{Introduction}
\label{sec:intro}
Digital video compression has been studied for over 50 years. It is a challenging research topic to exploit both spatial and temporal redundancies inside the video data. The concept of using motion compensation to reduce temporal correlation for video coding first appeared in 1969 \cite{rocca1969}. Since then, motion estimation and coding have become indispensable components in a video coding system. Two critical components in a mainstream video codec are motion coding (including motion estimation and compensation) and residual image coding. Motion coding is used to reduce {\it temporal} redundancy, and residual coding is used to reduce {\it spatial} redundancy. This structure is thus often called {\it hybrid coding}. The influential and widespread international video standards in the past three decades, MPEG-2, AVC/H.264, HEVC/H.265, and VVC/H.266 all adopt this basic hybrid coding structure, although the fine details vary in different versions of standards. These standards specify three types of coding frames inside a Group of Pictures (GOP): I-frame (intra-coded), P-frame (predictive), and B-frame (bidirectional predictive). The P-frame coding process uses the previously coded frame to predict the target frame, and the B-frame coding uses two reference frames (often previous and future frames) to predict the target frame. In this paper, we focus on learning-based B-frame video coding.\\     
\indent
In the past few years, deep-learning techniques have been used in video compression. Up to now, most learned codecs adopt the hybrid coding structure of the classical coding systems; that is, it contains two major components: motion coding and residual image coding. It is generally believed that accurate motion compensation is a very effective way to reduce the temporal redundancy in the video. Only the remaining unpredictable (`new') pixels are coded using image coding techniques. Describing accurately the motion field around arbitrary shape objects often needs a large number of bits. For example, the HEVC standard defines a variety of block partitions to specify regions sharing the same motion vectors \cite{Sullivan2012hevc}. \\
\indent
Thanks to the advancement of neural networks, more accurate video predictors without transmitting bits are now available. Then, we need only to send the unpredictable pixels. Often the locations of unpredictable pixels are sparse. It costs many bits to send the precise location information. Hence, we develop a bootstrap strategy. Instead of transmitting motion or location information or both, we send the unpredictable pixel information in two layers. The \textit{base layer} sends the downsampled unpredictable information (containing locations and pixel values) to the decoder. This piece of information serves two purposes. It provides a rough, downsampled image of unpredictable pixels and contains information indicating which pixels are unpredictable. With a well-designed neural network, we generate a weighting map that merges predictable and unpredictable pixels to construct a good-quality target frame. Then, at the \textit{enhancement layer}, we send additional information (bits) to improve the quality of the final coded image.\\
\indent
Motivated by the above observations, we propose a learned video compression scheme without a motion coding module. It contains two image coding layers: the base and enhancement layers. The base layer consists of a video frame interpolator, a downsampling network, a neural network-based image compressor, and a super-resolution network (SR-Net). We adopt the efficient Conditional Augmented Normalization Flows (CANF) \cite{ho2022canf} for the image compressors at the base and enhancement layers. The frame interpolator produces the conditioning image for the base-layer CANF. The SR-Net upsamples the decoded base-layer image to recover a full-resolution image.
The enhancement layer consists of a multi-frame merging network, skip-mask generator, skip-mode coding module and CANF compressor. The multi-frame merging network combines all the image information available at both the encoder and the decoder to form a \textit{merged} image. The merged image serves as the conditioning signal for the enhancement-layer CANF. To this end, we design a merging map (weights) generator, a neural network accepting inputs from the upsampled base-layer image, and {two motion-warped reference frames}.
To improve the coding efficiency of the enhancement-layer compressor, we design a skip-mode coding technique. A neural network generates a binary skip mask {${SM}_t$} according to the predicted motion information, the base-layer merged output, and the enhancement-layer hyperprior output. The skip mask specifies the locations of significant and insignificant latent samples. The insignificant samples are skipped from coding; at the decoder, they are replaced by the corresponding mean values predicted by the enhancement-layer hyperprior module. The detailed skip-mode coding operation is described in the supplementary document.\\
\indent
Our contributions are summarized as follows.
 \vspace{-0.2cm}
\begin{itemize}
    \item We propose a two-layer B-frame coding framework that skips motion information from coding. 
 \vspace{-0.2cm}
    \item We introduce a multi-frame merging network to combine the base-layer and enhancement-layer frames in constructing a high-quality predictor for the enhancement-layer CANF compressor.
 \vspace{-0.2cm}
\end{itemize}
\vspace{-0.1cm}
 We implement the above ideas in an end-to-end learned B-frame video compression system. Because the input image to the base-layer compressor has a much smaller dimension, our system has much lower computational complexity (about 45\% lower in terms of encoding MACs) than B-CANF\cite{chen2022bcanf}, a typical hybrid coding system with similar coding components. 

\begin{figure*}[t]
\vspace{-0.45cm}
\begin{center}
\includegraphics[width=0.8\textwidth]{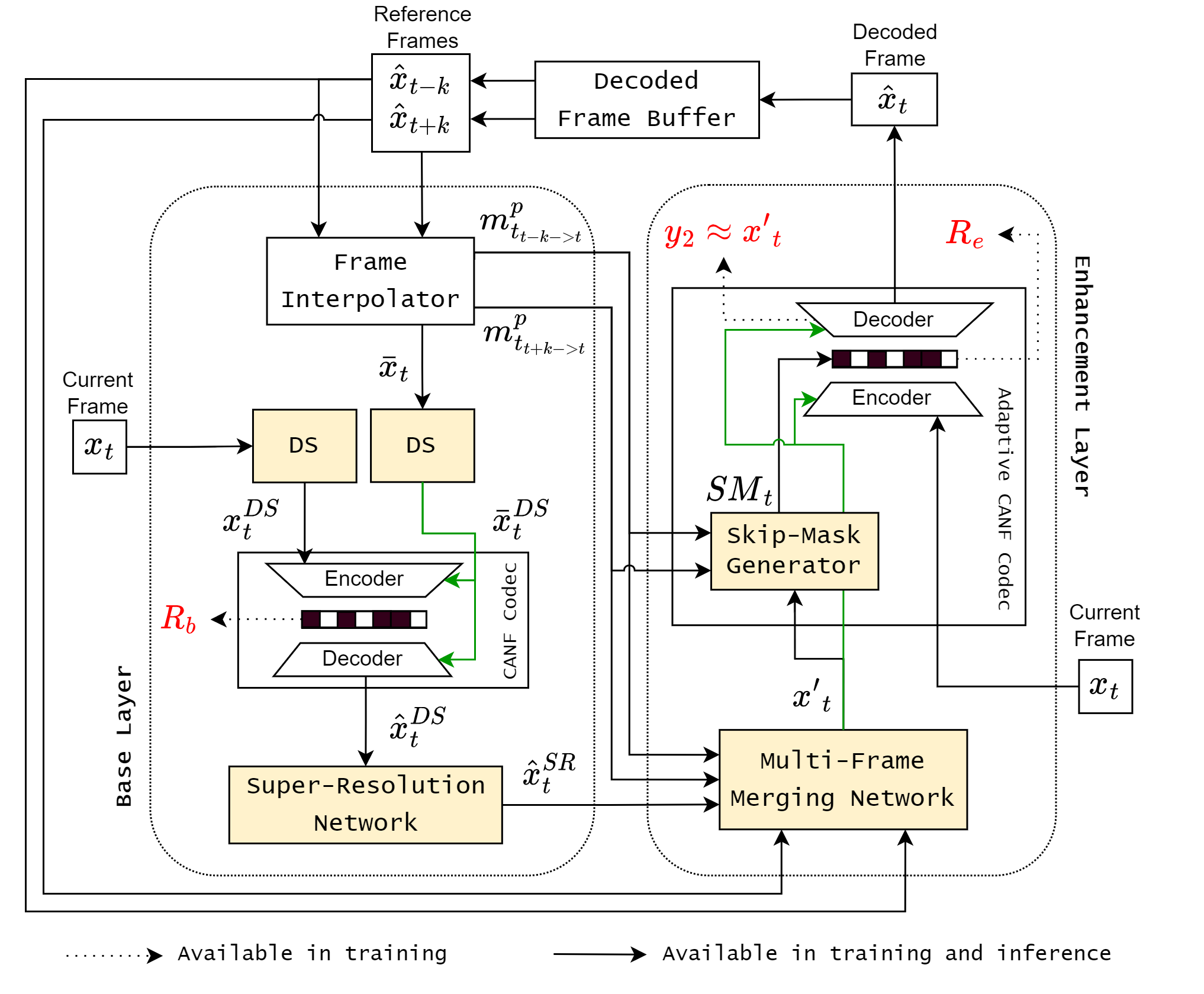}
\end{center}
\vspace{-0.5cm}
   \caption{The proposed two-layer conditional B-frame coding system without motion coding. It includes a low-resolution CANF compressor and a full-resolution adaptive CANF compressor. {The input frame $x_t$ is encoded based on its reference frames $\hat{x}_{t-k}$, $\hat{x}_{t+k}$, with the decoded frame $\hat{x}_t$ representing a lossy reconstruction of $x_t$}. {The yellow blocks denote our proposed components. 
   The green solid lines represent the conditioning signals for the CANF compressors. The red symbols are available only in the training phase.}}
\label{fig:framework}
\vspace{-0.5cm}
\end{figure*}

\section{Related Works}
\label{sec:related-works}
\subsection{Deep Video Compression}
Most existing deep video compression schemes adopt the hybrid coding structure with motion and residual coding, and focus on P-frame coding. For example, an early work by Lu et al. \cite{lu2019dvc, lu2020end} presents an efficient deep video coding scheme that replaces nearly all the key components in the classical coding architecture by deep neural networks. Recent deep video compression papers often use similar hybrid coding structures and focus on improving various components, e.g.~motion coding~\cite{hu2020improving, agustsson2020scale, lin2020m}, residual coding~\cite{feng2020learned}, feature-space coding~\cite{hu2021fvc}, content-adaptive coding~\cite{lu2020content}, coding mode prediction \cite{hu2022coarse}, and contextual coding~\cite{li2021deep, li2022hybrid, ho2022canf}. One notable trend is the use of conditional/contextual coding to replace traditional residual coding. For example, the contextual coding papers~\cite{li2021deep, li2022hybrid,ho2022canf} adopt this concept to achieve high coding performance. \\ 
\indent
There are only a few attempts at eliminating motion coding (i.e.~not transmitting motion information) in a video codec \cite{cheng2019learning}, \cite{zou2021learned,zou2021adaptation}, \cite{chen2022learning}.
Both Cheng et al. \cite{cheng2019learning} and Zou et al. \cite{zou2021learned,zou2021adaptation} adopt the hierarchical B-frame GOP structure, where Cheng et al. \cite{cheng2019learning} encode frame differences. They adjust the temporal distance in calculating frame differences based on the motion characteristics. 
Zou et al. \cite{zou2021learned,zou2021adaptation} compute the pyramid features of the reference and target frames and derive motion information from the transmitted features at the decoder.
On the other hand, Chen et al. \cite{chen2022learning} focus on P-frame coding and transmit the displaced frame differences instead of sending motion information. 
This concept of video compression without motion coding received little attention. The reason may be because of its inferior coding performance, although employing only one compressor significantly reduces the complexity. 
As discussed earlier, we observe that a low-rate base layer is needed to efficiently convey unpredictable pixels to improve compression performance. \\
\indent
Up to now, only a handful of deep video compression schemes address B-frame coding. In addition to \cite{cheng2019learning,zou2021learned,zou2021adaptation}, Wu et al. \cite{wu2018video} introduce an early deep video compression system that encodes B-frames hierarchically using a simple image interpolation method. Often, the motion information for the two reference frames are coded and transmitted \cite{djelouah2019neural},\cite{yang2020Learning}. 
Pourreza et al. \cite{pourreza2021extending} extend the P-frame coding method to encode B-frames using only one motion field. Yilmaz et al \cite{yilmaz2021end} propose learned hierarchical bi-directional
video compression (LHBDC) that employs a temporal motion vector predictor to reduce the motion bitrate. It produces impressive coding performance when compared to the prior learned P-frame and B-frame codecs. This scheme was refined and extended to flexible rate compression by \c{C}etin et al. \cite{ccetin2022flexible}. 

\subsection{CANF Compressors}
Ho et al. \cite{ho2022canf} propose Conditional Augmented Normalizing Flows (CANF) \cite{ho2022canf} by combining the concept of conditional coding with an efficient deep image compression architecture, Augmented Normalizing Flows (ANF) \cite{ho2021anfic}. In theory, conditional coding is more efficient than the residual coding that has been used in typical hybrid coding systems \cite{ladune2021conditional},\cite{ho2022canf}. Therefore, several conditional coding structures \cite{li2021deep}, \cite{li2022hybrid},\cite{ho2022canf} are proposed, showing promising compression performance. CANF can replace the usual VAE compressors in the hybrid coding structure and produce the state-of-the-art performance in P-frame coding \cite{ho2022canf}. 
Recently, Chen et al. \cite{chen2022bcanf} apply CANF to B-frame coding. They still use the hybrid coding structure and show the state-of-the-art performance with an additional frame-adaptive coding technique. We thus also use CANF as the image compressor in our system, but we do not use the hybrid coding structure. \textcolor{black}{Our adaptive CANF differs from the basic CANF (shown in the supplementary document) in that it incorporates a skip-mask generator and a skip-coding mechanism}. 

\section{Proposed Method}
\subsection{System Overview}
Figure \ref{fig:framework} shows our two-layer conditional coding scheme for B-frame video compression. The basic building block of our system is CANF \cite{ho2022canf}, and our intra-coding is an ANF image compressor from \cite{ho2021anfic}. Our framework has two coding layers: a low-resolution CANF compressor (the base layer) and a full-resolution CANF compressor (the enhancement layer).

\subsection{Base Layer}
The base layer comprises a frame interpolator, downsampler (DS), super-resolution network (SR-Net), and CANF compressor.
We adopt an off-the-shelf high-performance video interpolator, RIFE \cite{huang2022real}, as our frame interpolator.

\begin{figure} [tbp]
\begin{center}
\centering
\includegraphics[width=0.48\textwidth]{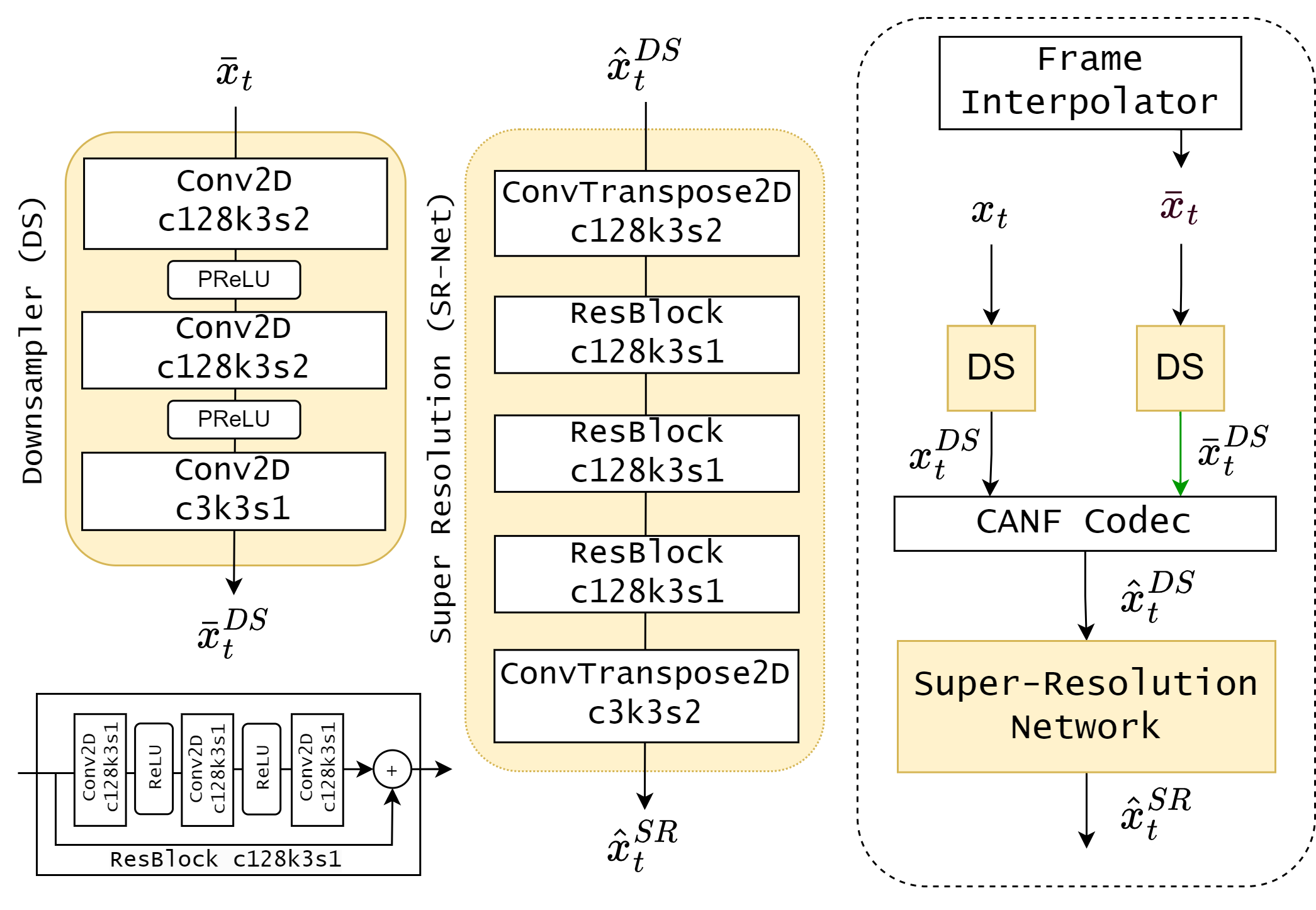}
\end{center}
\vspace{-0.45cm}
   \caption{Illustration of the base-layer components. The yellow blocks indicate our proposed modules, i.e.~\textcolor{black}{the downsampler (DS) and super-resolution network (SR-Net)}. The high-resolution input image $\bar{x}_t$ to the downsampler is produced by the RIFE frame interpolator. It is downsampled by \textcolor{black}{a factor of 4}, with the resulting signal $\bar{x}_t^{DS}$ serving as the conditioning signal for the low-resolution CANF compressor, which encodes the downsampled version $x_t^{DS}$ of the target frame $x_t$. The compressor output $\hat{x}_t^{DS}$ is upsampled by SR-Net as $\hat{x}_t^{SR}$.}
\label{fig:base-layer-fig}
\vspace{-0.65cm}
\end{figure}

As shown in Figure~\ref{fig:base-layer-fig}, the downsampling network (DS) downsamples the pixel-domain interpolated frame $\bar{x}_t \in R^{3 \times H \times W}$ to $\bar{x}^{DS}_t \in R^{3 \times H/4 \times W/4}$, where $W,H$ are the width and height of the target frame $x_t \in R^{3 \times H \times W}$, respectively. The same DS is also applied to $x_t$ to produce ${x}^{DS}_t \in R^{3 \times H/4 \times W/4}$. 
The \textcolor{black}{downsampled interpolated frame $\bar{x}^{DS}_t$ then serves as the conditioning signal for the CANF compressor to compress the downsampled target frame ${x}^{DS}_t$}. After the compression step, we recover the resolution of the coded downsampled target frame \textcolor{black}{$\hat{x}^{DS}_t\in R^{3 \times H/4 \times W/4}$} to its original resolution by a super-resolution network (SR-net). 

\textbf{Downsampling Network (DS)}. The DS network is composed of convolutional layers \textcolor{black}{and residual blocks} (Figure \ref{fig:base-layer-fig}). Specifically, we use two convolutional layers with stride 2 to achieve a downsampling factor of $m=4$. 

\textbf{Base-Layer CANF Compressor}. The base-layer CANF compressor encodes the downsampled target frame \textcolor{black}{${x}^{DS}_t$} by taking the \textcolor{black}{downsampled version $\bar{x}^{DS}_t$} of the interpolated frame $\bar{x}_t$ as a conditioning signal. Note that the base-layer CANF compressor is to be distinguished from another CANF compressor in the enhancement layer.

\textbf{Super-Resolution Network (SR-Net)}. The SR-Net is to interpolate the low-resolution coded target frame \textcolor{black}{$\hat{x}^{DS}_t$ to its original resolution $\hat{x}^{SR}_t \in R^{3 \times H \times W}$}. We perform upsampling using transpose convolutions with stride 2. The network architecture is detailed in Figure \ref{fig:base-layer-fig}. 

\begin{figure} [tbp]
\begin{center}
\centering
\includegraphics[width=1\linewidth]{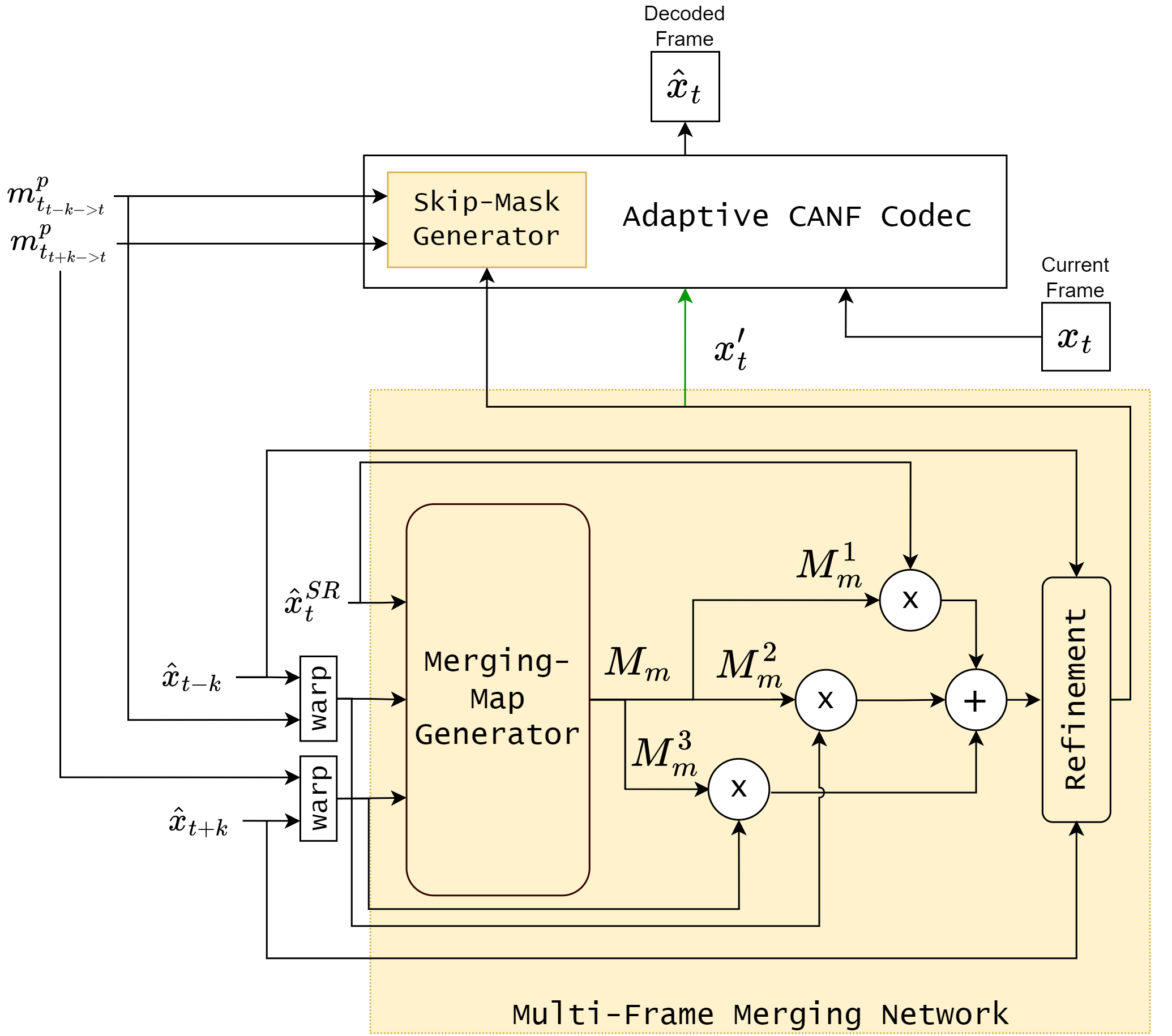}
\end{center}
\vspace{-0.45cm}
   \caption{\textcolor{black}{Illustration of the enhancement-layer components. The multi-frame merging network inside the enhancement layer is designed to combine $\hat{x}^{SR}_t$, $warp$(\textcolor{black}{$\hat{x}_{t-k}$, $m^p_{t_{t-k \rightarrow t}}$), and $warp$($\hat{x}_{t+k}$}, $m^p_{t_{t+k \rightarrow t}}$) to produce a refined merged output $x'_t$ according to a merging map generated by the merging-map generator.}}
\label{fig:enh-layer-fig}
\vspace*{-0.1in}
\end{figure}

\subsection{Enhancement Layer}
To obtain a high-quality output at the end of this stage, we introduce a multi-frame merging network. The network takes three inputs: the SR-Net output $\hat{x}^{SR}_t$ and the two warped (motion-compensated) reference frames $warp$($\hat{x}_{t-k}$, $m^p_{t_{t-k \rightarrow t}}$), $warp$($\hat{x}_{t+k}$, $m^p_{t_{t+k \rightarrow t}}$). It produces a floating-point weighting map ${M}_m \in R^{3 \times H \times W}$ with three normalized values for each sample, which are the weightings used to combine the three input frames. The weighted sum of these three input frames is further refined using a \textcolor{black}{refinement module (Refine-Net)} to generate the final output image $x'_t \in R^{3 \times H \times W}$. The architectures of the merging-map generator and Refine-Net are provided in the supplementary document.

We then use \textcolor{black}{the second CANF compressor operating in the original image resolution to produce a high-quality coded frame. The merged output $x'_t$ from the base layer serves as the conditioning signal to compress the target frame $x_t$. To reduce the bit consumption in arithmetic coding, we propose a skip-mode coding mechanism. The skip-mask generator network produces \textcolor{black}{a binary skip-mask ${SM}_t \in \{0,1\}^{128 \times H/16 \times W/16}$} that determines which latent samples are transmitted in the arithmetic coding process. We modify the CANF compressor from \cite{ho2021anfic} to work with ${SM}_t$. The same skip mask is used at the decoder to identify the locations of non-skipped samples decoded from the transmitted bitstream. The reconstructed frame $\hat{x}_t \in R^{3 \times H \times W}$ is stored in the decoded frame buffer and is used in the next coding cycle.}

\begin{figure*}[ht]
  \centering
  \begin{subfigure}{0.18\linewidth}
    \includegraphics[width=0.99\linewidth]{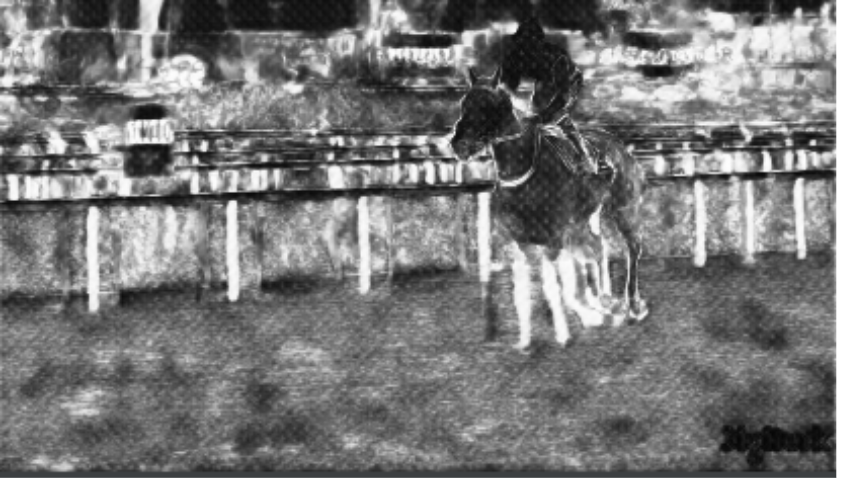}

    \label{fig:mask-a}
  \end{subfigure}
  \hfill
  \begin{subfigure}{0.18\linewidth}
    \includegraphics[width=0.99\linewidth]{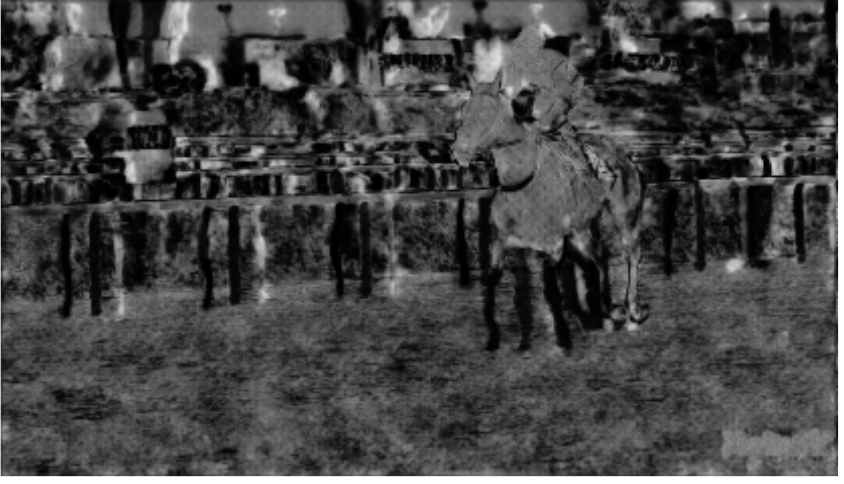}

    \label{fig:mask-b}
  \end{subfigure}
  \hfill
  \begin{subfigure}{0.18\linewidth}
    \includegraphics[width=0.99\linewidth]{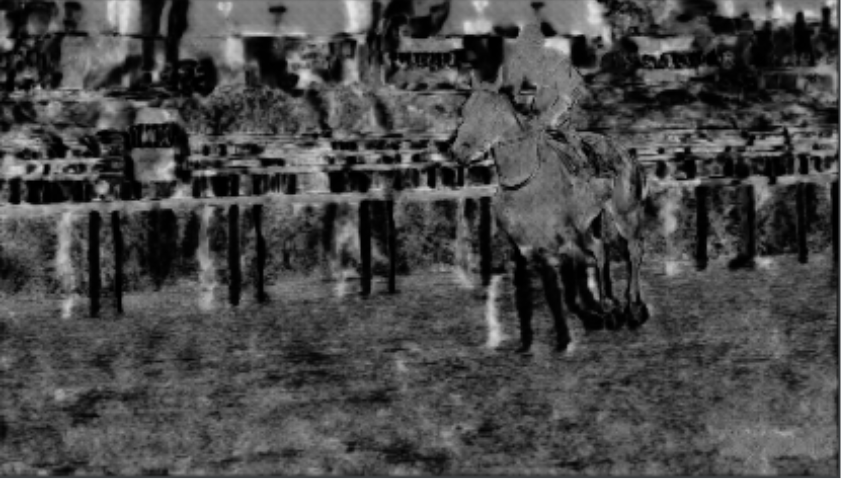}

    \label{fig:mask-c}
  \end{subfigure}
  \hfill
  \begin{subfigure}{0.18\linewidth}
    \includegraphics[width=0.99\linewidth]{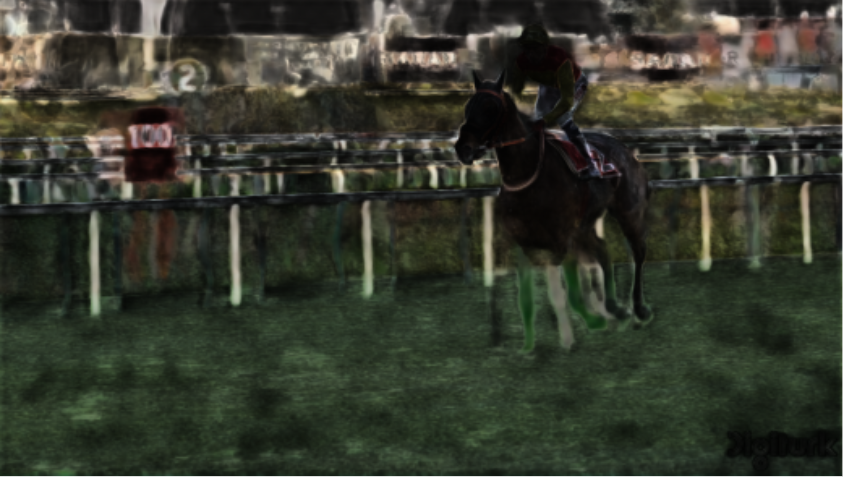}

    \label{fig:mask-d}
  \end{subfigure}
  \hfill
  \begin{subfigure}{0.18\linewidth}
    \includegraphics[width=0.99\linewidth]{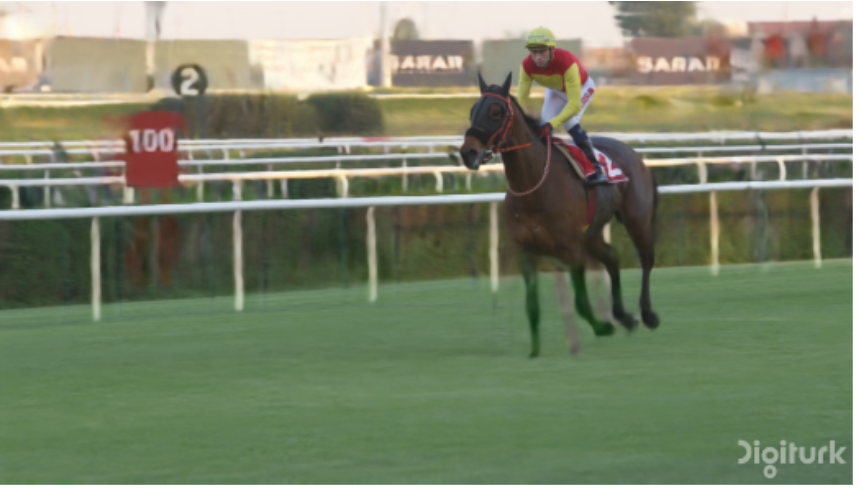}
    
    \label{fig:mask-e}
  \end{subfigure} \\
  \begin{subfigure}{0.18\linewidth}
    \includegraphics[width=0.99\linewidth]{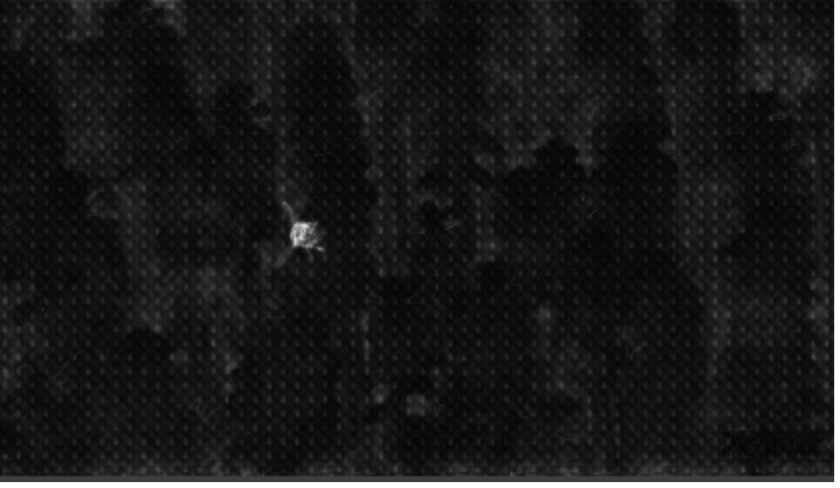}
    \caption{$M^1_m$}
    \label{fig:mask-f}
  \end{subfigure}
  \hfill
  \begin{subfigure}{0.18\linewidth}
    \includegraphics[width=0.99\linewidth]{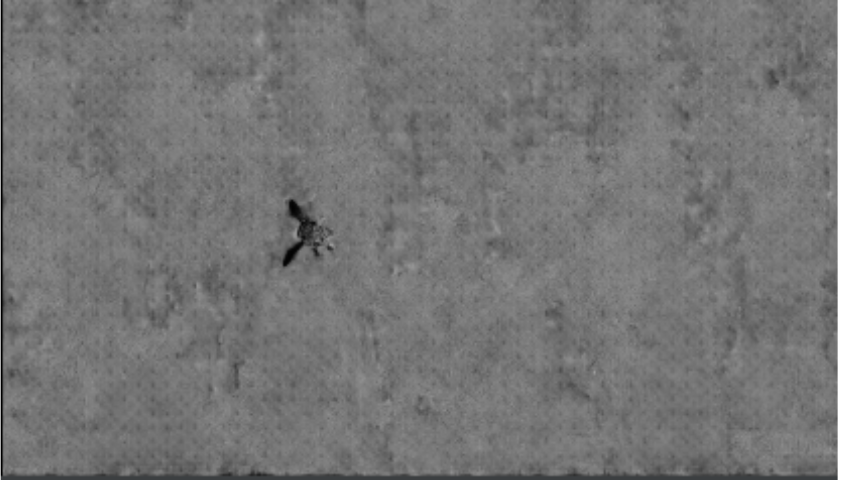}
    \caption{$M^2_m$}
    \label{fig:mask-g}
  \end{subfigure}
  \hfill
  \begin{subfigure}{0.18\linewidth}
    \includegraphics[width=0.99\linewidth]{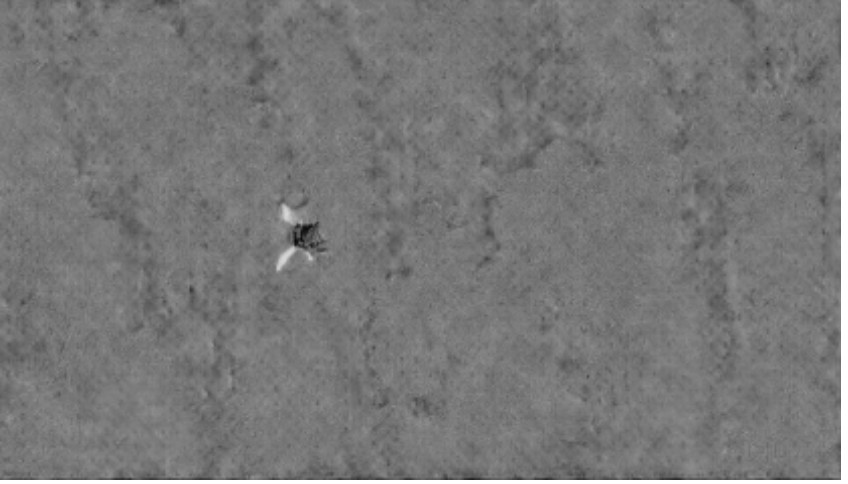}
    \caption{$M^3_m$}
    \label{fig:mask-h}
  \end{subfigure}
  \hfill
  \begin{subfigure}{0.18\linewidth}
    \includegraphics[width=0.99\linewidth]{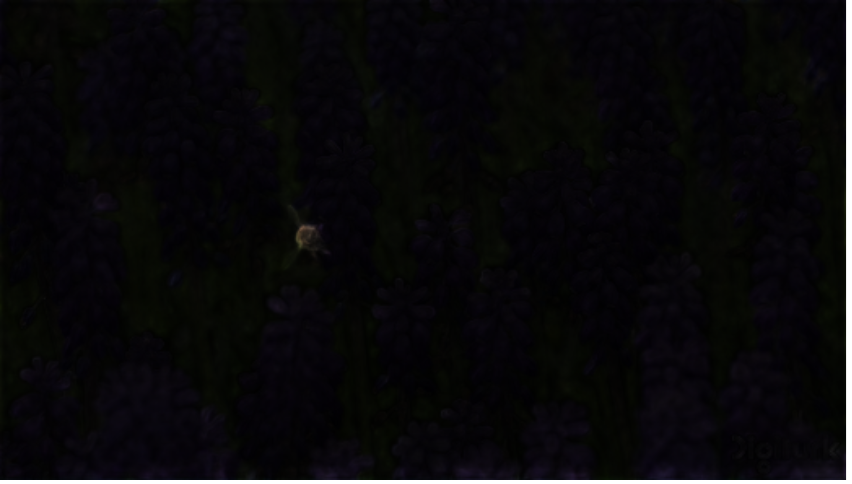}
    \caption{$M^1_m * \hat{x}^{SR}_t$}
    \label{fig:mask-i}
  \end{subfigure}
  \hfill
  \begin{subfigure}{0.18\linewidth}
    \includegraphics[width=0.99\linewidth]{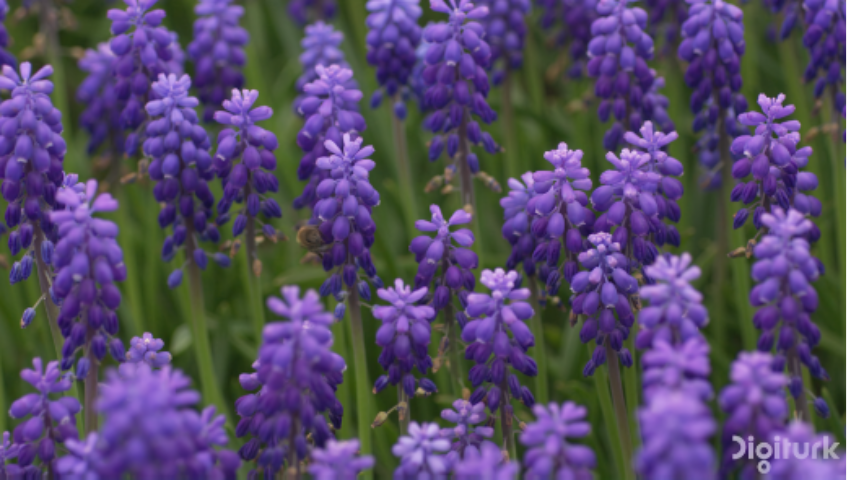}
    \caption{$x'_t$}
    \label{fig:mask-j}
  \end{subfigure}\\
    \vspace{-0.15cm}
  \caption{Visualization of intermediate results produced by multi-frame merging network (MFMN). The top row (Jockey) has fast moving background and the bottom row (HoneyBee) has slow moving background.}
      \vspace{-0.5cm}
  \label{fig:mfmn-mask-example}
\end{figure*}

\textbf{Multi-Frame Merging Network (MFMN)}. Inspired by \cite{yilmaz2021end}, we develop a multi-frame merging network, which produces the weighting maps used to combine the output $\hat{x}^{SR}_t$ of SR-Net and the two warped reference frames, $warp$(\textcolor{black}{$\hat{x}_{t-k}$, $m^p_{t_{t-k \rightarrow t}}$) and $warp$($\hat{x}_{t+k}$}, $m^p_{t_{t+k \rightarrow t}}$). \textcolor{black}{Therefore, the output channel number is three and the softmax operator is used for scaling.} Figure \ref{fig:mfmn-mask-example} illustrates the operations of our multi-frame merging network. In this example, the upper video is Jockey, and the lower one is HoneyBee. The three weighting maps $M_m^i \in R^{1 \times H \times W},i=1,2,3$, the weighted output $M_m^1*\hat{x}_t^{SR}$ of the coded base layer $\hat{x}^{SR}_t$, and the final combined output $x'_t$ are shown for one typical frame in these two sequences. The HoneyBee video is a slow-motion sequence; only a tiny honeybee has fast motion. Therefore, most of the background can be predicted well from the two reference frames. In contrast, both the object and background are moving in Jockey, and thus it is important to extract the locations of unpredictable pixels and their values from the decoded low-resolution image.

\begin{figure}[ht]
\begin{center}
\includegraphics[width=0.99\linewidth]{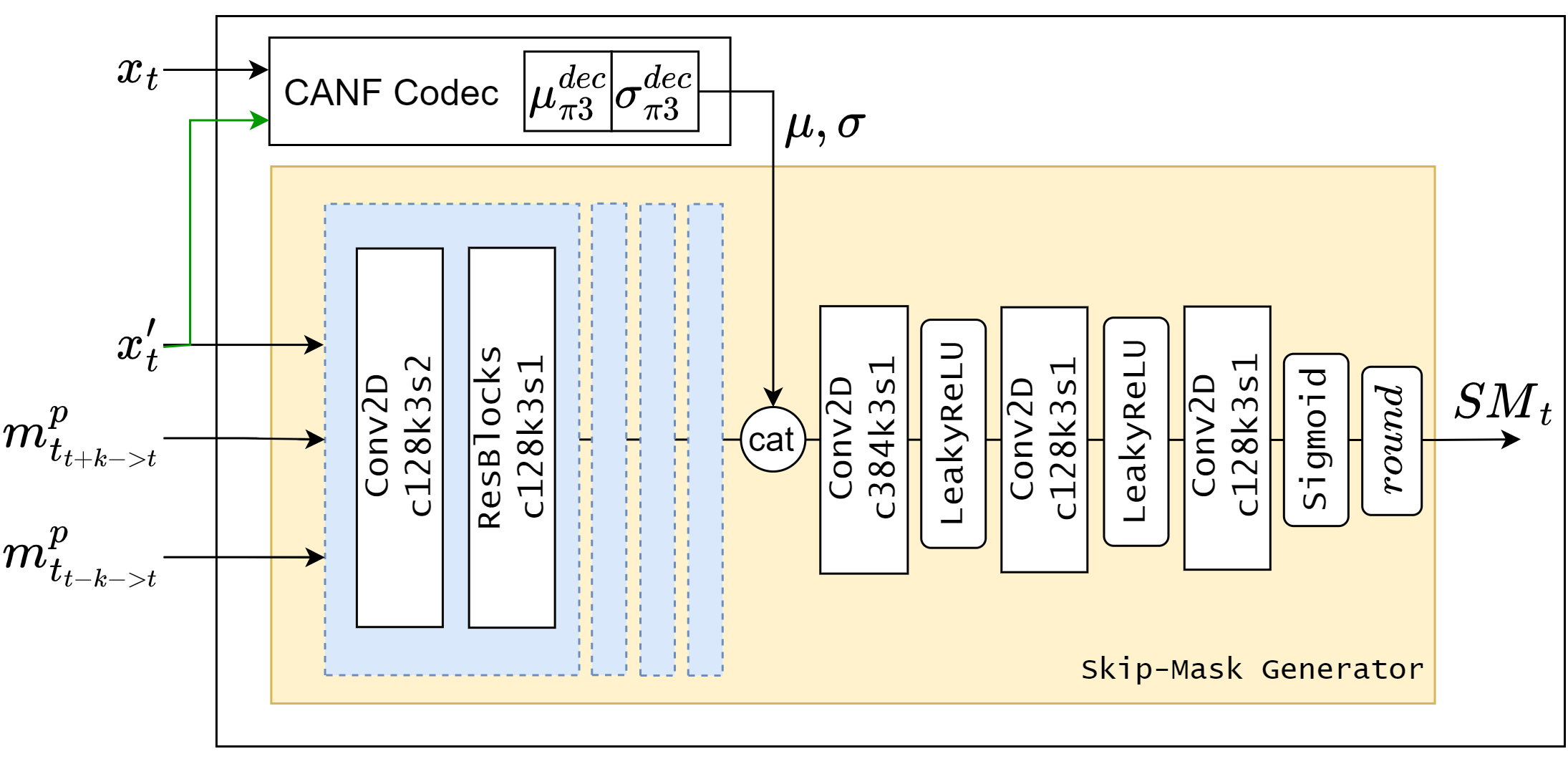}
    \label{fig:short-x}
  \hfill
\end{center}
\vspace{-0.45cm}
\caption{\textcolor{black}{The skip-mask generator consists of convolutional layers and residual blocks. We use a sigmoid function to scale the output to a range between 0 and 1, followed by using a rounding operator to create a binary map.}}
\label{fig:skip-mask-generator}
\vspace{-0.25cm}
\end{figure}

\begin{figure}[ht]
\begin{center}
    \begin{subfigure}{0.31\linewidth}
    \includegraphics[width=0.99\linewidth]{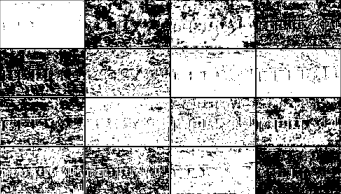}
    \caption{$\lambda$ = 256}
    \label{fig:short-a}
  \end{subfigure}
  \begin{subfigure}{0.31\linewidth}
    \includegraphics[width=0.99\linewidth]{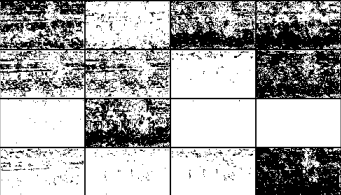}
    \caption{$\lambda$ = 512}
    \label{fig:short-b}
  \end{subfigure}
   \\
  \begin{subfigure}{0.31\linewidth}
    \includegraphics[width=0.99\linewidth]{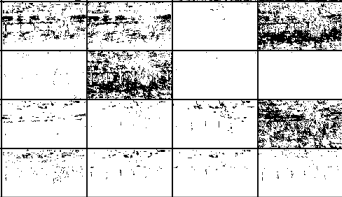}
    \caption{$\lambda$ = 1024}
    \label{fig:short-c}
  \end{subfigure}
  \begin{subfigure}{0.31\linewidth}
    \includegraphics[width=0.99\linewidth]{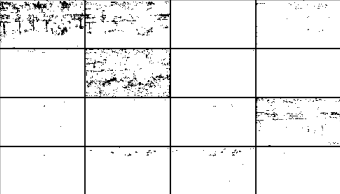}
    \caption{$\lambda$ = 2048}
    \label{fig:short-d}
  \end{subfigure}
\end{center}
\vspace{-0.45cm}
\caption{\textcolor{black}{Examples of the skip mask at different $\lambda$'s (smaller $\lambda$'s result in lower bitrates) for the model trained with mean-squared error (MSE). The transmitted latent variables have 128 channels; only 16 are shown for each bitrate.}}
\vspace{-0.55cm}
\label{fig:skip-mask-example}
\end{figure}

\textbf{Skip-mask Generation}. Our skip-mode coding mechanism has two main components: the (1) skip-mask generation and (2) skip-mode coding inside the arithmetic coder. The performance of the skip-mode coding largely relies on precise skip masks. Often the moving object boundaries and texture edges cannot be precisely predicted or upsampled from the low-resolution image. Hence, motion information provides clues to skip samples. Also, the decoded low-resolution image can provide object boundary and texture edge clues. Therefore, as shown in Figure \ref{fig:skip-mask-generator}, the first stage of our skip-mask generator takes inputs from the forward and backward motion fields, ${m}^p_{t-k \rightarrow t} \in R^{2 \times H \times W}$, ${m}^p_{t+k \rightarrow t}  \in R^{2 \times H \times W}$, and the merged image, $x'_t$. \textcolor{black}{We adopt the implementation of the skip-mask generation and skip-mode coding from \cite{alexandre2022two}.} \\
\indent
Furthermore, the skipped (not transmitted) samples are replaced by the mean values $\mu$ from the \textcolor{black}{hyperprior module} at the decoder. This operation is performed also at the encoder to reconstruct the decoded image. The mean $\mu$ and variance $\sigma$ produced by the hyperprior module also provide clues for skipping samples. Thus, the second stage of our skip-mask generator takes inputs from the hyperprior outputs, as shown in Figure \ref{fig:skip-mask-generator}.  Finally, a rounding operation is applied to generate the binary mask. We use the straight-through gradient strategy in training to solve the zero gradient problem caused by the round operator for mask binarization. Value 0 in the skip mask means skip mode, and value 1 means non-skip mode. We show a few masks in Figure \ref{fig:skip-mask-example}. Generally, more samples are skipped at lower bitrates.

\textbf{Adaptive CANF Compressor}.~Because our enhancement-layer CANF includes the skip-mode coding process described above, it is called Adaptive CANF. The details of our Adaptive CANF is described in the supplementary document.

\textbf{Frame-type Adaptive Coding}.~For better rate-distortion performance, {\it reference} B-frames should be coded with higher quality (at the cost of higher bitrates) than {\it non-reference} B-frames. 
To this end, we implement the frame-type adaptive (FA) coding proposed in \cite{chen2022bcanf}. Conceptually, the reference and non-reference B-frames are coded with two somewhat different models. This is achieved by applying a channel-wise affine transformation to the output features of every convolutional layer \textcolor{black}{in our CANF compressors}. 



\subsection{Training Procedure}
Our model is trained by using a multi-step training procedure. The hyper-parameters are chosen empirically. We use the ADAM \cite{kingma2014adam} optimizer with an initial learning rate of 1e-4. The batch size is set to 8. 
We train our model in four phases. Each phase has its own set of hyper-parameters and training loss functions. Some modules may be frozen during training; thus, only the other modules are trained in that step. 
Our training procedure is as follows.
\begin{enumerate}
\vspace{-0.15cm}
\item \textcolor{black}{We first train the frame interpolator (RIFE \cite{huang2022real}) with the initial model from \cite{huang2022real}. The loss function in this phase is $L = D(\bar{x}_t, x_t)$; that is, the output $\bar{x}_t$ of RIFE is trained to approximate the target frame $x_t$.
\item We train all the modules in the base layer in a few steps. The RIFE module is frozen at the beginning of this phase. First, we train only the downsampler and SR-Net (without the CANF compressor) using the loss function $L = D(\hat{x}^{SR}_t, {x}_t)$, where $\hat{x}^{SR}_t$ is the SR-Net output. When the first step reaches convergence, we include the CANF compressor between the downsampler and SR-Net in the second training step and the loss function is $L = D({\hat{x}^{DS}}_t, {{x}^{DS}_t}) + R_{b}$, where $R_{b}$ is the estimated coding bitrate of CANF in the base layer. Then, we train RIFE together with the downsampler, CANF, and SR-Net with $L = D({\hat{x}^{SR}_t}, {x_t}) + R_{b}$ to update the entire base layer. 
\vspace{-0.1cm}
\item After the base layer produces a target frame with reasonable quality, we proceed to train the enhancement layer. In this phase, we freeze the base layer.
We first train \textcolor{black}{the merging-map generator and refinement module inside MFMN with $L = D({x}'_t, x_t)$.}
Then, we train the MFMN together with the enhancement-layer CANF compressor without the skip-mask generator network. The loss function is $L = D(\hat{x}_t, x_t) + R_{b} + R_{e}$, where $R_{e}$ is the estimated coding bitrate of the enhancement-layer CANF. When the above training step converges, we include the skip-mask generator network and activate the skip-mode coding process inside the adaptive CANF compressor for training.
\item In the final phase, we train all the modules in an end-to-end manner. We append $Aux$ at the end of the loss function and introduce a parameter $\varepsilon$ in front of $R_{b}$. \textcolor{black}{$Aux$ refers to $(D(y_2, x'_t) + D(x'_t, x_t) + D(\hat{x}^{SR}_t, x_t)) * 0.01 * \lambda$. It functions as a regularizer for $y_2$, $x'_t$, and $\hat{x}^{SR}_t$. $y_2$ is the approximation of conditioning signal $x'_t$} from CANF codec \cite{chen2022bcanf}. The parameter 0.01 is recommended by \cite{ho2022canf} although our terms are slightly different. Thus, the final loss function is $L = D(\hat{x}_t, x_t) * \lambda + \varepsilon * R_{b} + R_{e} + Aux$.}
\end{enumerate}

In total, we use five epochs to train RIFE with the initial model from \cite{huang2022real}, five epochs to train the base layer, five epochs to train the enhancement layer, and 25 epochs to train all the modules in an end-to-end manner. We reduce the learning rate when the loss function reaches a plateau. To obtain models for different bitrates, we choose $\lambda$ = 256, 512, 1024, 2048 for training the mean-squared error (MSE) model and $\lambda$ = 4, 8, 16, 32 for training the multi-scale structural similarity index (MS-SSIM) model. The MSE model adopts MSE as the distortion measure $D(\cdot,\cdot)$, and the MS-SSIM model adopts MS-SSIM. The $\varepsilon$ parameter controls the bitrate (and thus image quality) of the base layer. In our experiment, $\varepsilon$ = 4 is chosen empirically. To generate different rate points, we first train the model for the highest rate point ($\lambda = 2048$) using the complete training procedure and then fine-tune (phase 4 only) the resulting model for the other rates for five epochs.


\section{Experiments}

\subsection{Dataset}
The Vimeo90K septuplet dataset \cite{xue2019video} was used to train our proposed method. It contains 91,701 7-frame sequences of resolution 448x256. During training, we randomly crop each frame to 256x256 and flip it horizontally and vertically. We evaluate our training models using the popular video coding test datasets: UVG \cite{mercat2020uvg} (7 videos) and HEVC Class B \cite{bossen2013common} (5 videos). The performance metrics are peak-signal-to-noise ratio (PSNR) and multi-scale structural similarity index (MS-SSIM) at several coding bitrates. We also calculate the BD-rate savings \cite{Bjontegaard2001}.

\begin{figure*}[h]
  \centering
  \begin{subfigure}{1\linewidth}
      \includegraphics[width=1\linewidth]{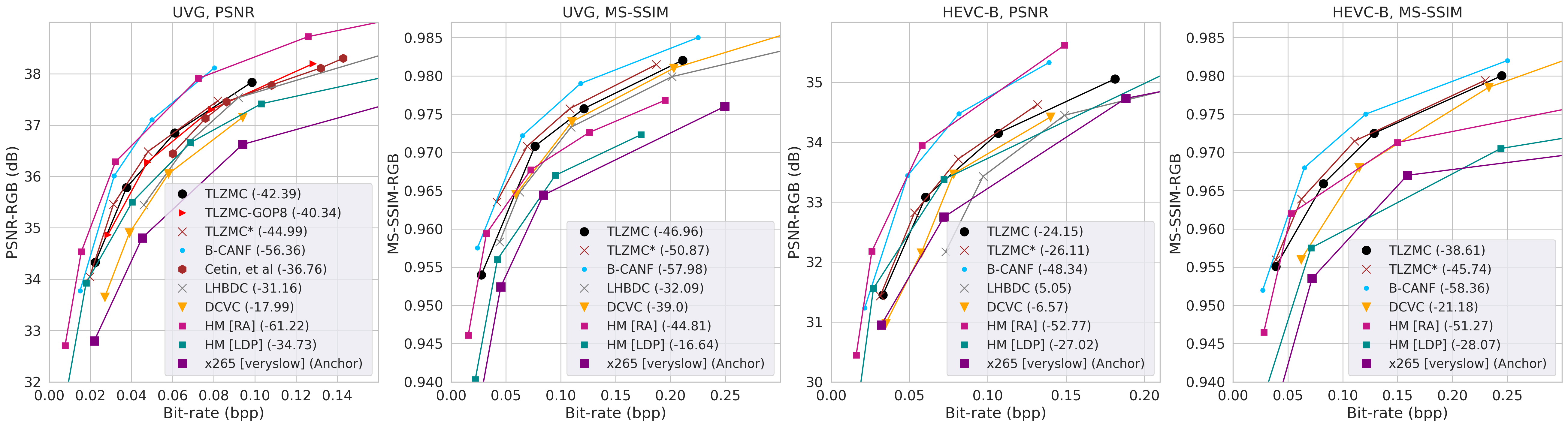}
  \end{subfigure}
   \vspace{-0.45cm}
  \caption{RD results (GOP=32) on UVG and HEVC Class B datasets measured in PSNR and MS-SSIM over bitrate (bpp). There are different evaluation settings for B-CANF (intra period=32, GOP=16), \c{C}etin et al. \cite{ccetin2022flexible} (GOP=16), and LHBDC (GOP=8). }
  \label{fig:experimental-results}
  \vspace{-0.45cm}
\end{figure*}

\subsection{Rate-Distortion (RD) Performance}

Figure \ref{fig:experimental-results} shows the RD performance on the test datasets using GOP=32. Our proposed method is denoted as TLZMC (Two-Layer Zero Motion Coding). When the FA technique is used, our method is denoted as TLZMC*. More results with different downsampling and super-resolution methods are provided in the supplementary document. Clearly, the RD performance of TLZMC* is somewhat better than that of TLZMC. Our methods are compared with DCVC \cite{li2021deep} (a conditional P-frame coding scheme) and the other hierarchical B-frame coding methods: LHBDC \cite{yilmaz2021end} (GOP=8), \c{C}etin et al. \cite{ccetin2022flexible} (GOP=16), and B-CANF \cite{chen2022bcanf} (hybrid-based coding with intra period=32 and GOP=16), which is the state-of-the-art B-frame coding scheme. For classical coding, we include the RD curves of HM 16.23 \cite{HM1623} with \textit{encoder\_lowdelay\_P\_main} configuration (LDP) and with \textit{encoder\_randomaccess\_main} configuration (RA / Random Access) and x265 \cite{FFMPEG} with \textit{veryslow} mode (zerolatency). The BD-rate saving in the parenthesis is calculated using x265 (veryslow) with GOP=32 as anchor. We perform coding on all available frames for the UVG dataset, but only on the first 100 frames for the HEVC Class B dataset.

Except for B-CANF, \textcolor{black}{our method} outperforms all the other deep video codecs in PSNR. It should be noted that the B-CANF performance is based on an intra period of 32 and a GOP of 16 using its \textit{B*-frame} technique. In comparison with the classical codecs, \textcolor{black}{our method} outperforms HM (LDP) and x265 (veryslow) but is lower than HM (RA). Regarding MS-SSIM, \textcolor{black}{our method} is slightly lower than B-CANF but outperforms the other deep video codecs and the classical codecs. It is interesting to observe that the performance of our method is closer to that of B-CANF at lower bitrates in MSE and MS-SSIM models.

\begin{table}[h]
\centering
\begin{tabular}{ p{1.0cm} c c c  }
 \hline
 
 $\lambda$ & {TLZMC} & {TLZMC*}\\
  \cline{2-3}
& R / NR / AVG & R / NR / AVG\\
 \hline 
 256 &  5.38 / 8.71 / 6.80\% & 18.05 / 15.96 / 17.41\%\\
 512 & 4.30 / 7.55 / 5.53\% & 11.58 / 8.45 / 10.52\%\\
 1024 & 2.99 / 6.43 / 4.51\% & 8.08 / 5.16 / 7.01\%\\	 
 2048 & 2.55 / 5.29 / 3.46\% & 5.31 / 2.83 / 4.31\%\\ 
 
 \hline 
 Average & 3.80 / 6.99 / 5.07\%  & 10.75 / 8.10 / 9.81\%\\
 \hline
\end{tabular}

 \caption{\textcolor{black}{Percentages of the base-layer bit rate for 100 frames in all videos in the HEVC-B dataset. The percentages of the enhancement-layer bitrate can be derived by (100 - BL)\%. The total bitrate excludes intra frames. R: reference frames, NR: non-reference frames, and \textcolor{black}{AVG: the average percentages of the base-layer bitrate over both reference and non-reference frames.}}}
\vspace{-0.65cm}
 \label{tab:bl-el-distribution}
\end{table} 

Table \ref{tab:bl-el-distribution} presents the bit distribution between the base and enhancement layers. Generally, the base layer consumes less than 7\% of the total bitrate in average. However, when employing frame-type adaptive coding, the base-layer bitrate exhibits increased flexibility, reaching up to 18\% in reference frames and 16\% in non-reference frames.

\subsection{Skip-Mask and Skip-Mask Generator}
We show the benefit of our skip mask by calculating the percentage of retained latent samples (transmitted samples) at various bitrates ($\lambda$ values). Table \ref{tab:skip-mask-percentage} shows the statistics of retained samples using the MSE model on the UVG dataset (GOP=8). The percentage of retained samples is lower at lower bitrates (lower $\lambda$ values) because fewer bits can be used to send transmitted samples. The average retained rates on the UVG dataset for $\lambda$ = 256, 512, 1024, and 2048 are 28.28\%, 36.23\%, 57.17\%, and 69.93\%, respectively.

\begin{table}[b]
  \vspace{-0.45cm}
\centering
\begin{tabular}{ p{2.5cm} c c c c }
 \cline{1-5}
 \multirow{2}{*}{Sequence}& \multicolumn{4}{c}{$\lambda$} \\
\cline{2-5}
   & 256 & 512 & 1024 & 2048  \\
 \hline 

 YachtRide & 40.75 & 53.21 & 64.01& 70.22\\
 Bosphorus & 23.79 & 34.01 & 43.21 & 58.63\\
 ShakeNDry & 29.10 & 39.43 & 55.64& 75.77\\
 ReadySteadyGo & 39.65 & 44.23 & 56.16& 64.85\\
 Beauty & 20.99 & 29.73& 81.63& 93.65\\
 Jockey & 35.03 & 38.77 &65.13 & 67.45\\
 HoneyBee & 8.66 & 14.21 & 34.42& 58.96\\
 \hline 
 Average & 28.28  & 36.23  & 57.17 & 69.93 \\
 \hline
\end{tabular}

 \caption{The percentages of retained (transmitted) latent samples at different $\lambda$ values. Smaller $\lambda$'s result in lower bitrates.}
 \label{tab:skip-mask-percentage}
  \vspace{-0.15cm}
\end{table} 

The skip-mask generator has two sets of inputs: (1) predicted motion data and merged frames, and (2) $\mu$ and $\sigma$ from the hyperprior. Table \ref{tab:bd-rate-skip-map} shows how each input contributes to the BD-rate saving. The evaluation is performed on the UVG dataset with GOP=32, and each model is separately trained in an end-to-end manner (phase 4). The BD-rate saving of both sets of inputs is significantly better that of the individual sets alone.

\begin{table}[t]
  \vspace{-0.1cm}
\centering
\begin{tabular}{ c p{3.5cm} c }
 \hline
  & Input & BD-Rate Saving \\
 \hline 
  (i) & Without skip mask & -35.87\\
  (ii) & $x'_t$, $m^p_{t_{t-k \rightarrow t}}$, $m^p_{t_{t+k \rightarrow t}}$ & -38.12\\
  (iii) & $\mu$, $\sigma$ & -39.74\\
  (iv) & (ii) \& (iii) & -42.39 \\
 \hline 
\end{tabular}

 \caption{Ablation study of the inputs to the skip-mask generator tested on the UVG dataset (GOP=32).}
 \label{tab:bd-rate-skip-map}
 \vspace{-0.6cm}
\end{table}


\subsection{GOP Size}
To better understand the RD performance under different GOP settings, we include our RD performance on the UVG dataset using GOP=8 (TLZMC-GOP8) in Figure \ref{fig:experimental-results}. As shown, a larger GOP size leads to a slightly higher BD-rate saving.  When tested on the UVG dataset, \textcolor{black}{our method} with GOP=32 performs comparably to B-CANF at lower bitrates in terms of both PSNR and MS-SSIM.


\subsection{Computational Complexity}
\label{sec:computational-complexity}
The complexity of our method is shown in Table \ref{tab:model-complexity} in terms of model size, runtimes, and multiply-and-accumulate operations (MACs). The test is run on GTX 2080Ti with GOP=32 on the UVG dataset. The MACs is calculated when encoding the first B-frame in a GOP. The encoding and decoding runtimes are averaged on the first 100 frames of Beauty sequence (UVG dataset), following the setting of \cite{chen2022bcanf}. Our MAC number is extracted using PyTorch library fvcore \cite{fvcore}. Because of the use of the CANF compressor in the base and enhancement layers, our model size reaches 39.9M, which is approximately 1.5x larger than the others. However, the number of pixels to the base-layer compressor is one-sixteenth of the full image resolution, resulting in a significant reduction in MACs and runtimes. Particularly, our encoder has only a slightly larger amount of computation than the decoder, while the other schemes have much higher encoder computation. This is because our encoder does not need to perform extra motion estimation for motion coding. 
Notably, our method has the lowest encoding time and its decoding time is very close to that of LHBDC, which has the lowest decoding time. Our encoding and decoding MACs are also very competitive.

We present a breakdown analysis of the encoder's model size and MACs in Table \ref{tab:model-complexity-breakdown}. Clearly, the base-layer multi-frame merging network and the enhancement-layer adaptive CANF use more than 80\% of calculations. They may be subjected to further study for reducing computation.


\section{Conclusion}
We propose a two-layer video compression framework without motion coding. It is different from the mainstream hybrid-based coding framework in which motion coding is an essential component. 

\begin{table}[t]
  \vspace{-0.1cm}
  \centering
\begin{tabular}{ p{1.3cm} p{0.7cm} p{0.7cm} p{1.2cm} p{0.7cm} p{1.2cm} }
\hline
  \multirow{2}{*}{Model} & \multirow{2}{*}{Size} & \multicolumn{2}{c}{Encode} & \multicolumn{2}{c}{Decode}  \\
 \cline{3-6} 
 & & Time & MACs & Time & MACs \\
 \hline
 DCVC & 8M & 7.70s & 1.05M/px & 28.97s & 0.68M/px \\
 LHBDC & 23.5M & 1.19s & 1.94M/px & 0.73s & 1.12M/px\\
 B-CANF & 24M & 1.69s & 2.70M/px & 1.09s & 1.97M/px\\
 TLZMC & 39.9M &  \textbf{0.87s} & \textbf{1.50M/px} & \textbf{0.76s} & \textbf{1.45M/px} \\
  \hline
\end{tabular}
 \vspace{-0.15cm}
 \caption{Computational complexity comparison with DCVC \cite{li2021deep} (P-frame coding), LHBDC \cite{yilmaz2021end} and B-CANF \cite{chen2022bcanf}.}

 \label{tab:model-complexity}
 \vspace{-0.25cm}
\end{table}

\begin{table}[t]
  \centering

\begin{tabular}{ p{1.6cm} c c c c  }

\hline
  Modules & Size &  Ratio & MACs &  Ratio \\
 \hline
 \multicolumn{3}{p{3 cm}}{\textit{Frame Interpolator}} & &\\

 RIFE & 10.7M & 26.94\% & 0.09M/px & 5.76\% \\
  \hline
 \multicolumn{3}{p{2 cm}}{\textit{Base Layer}} & & \\

 CANF & 12.6M & 31.61\% & 0.06M/px & 4.16\%\\
 DS & 0.1M & 0.01\% & 0.01M/px & 0.01\%\\
 SR-Net & 0.3M & 0.76\% & 0.09M/px & 6.63\%\\
 MFMN & 1.5M & 3.74\% & 0.43M/px & 29.03\%\\
  \hline
\multicolumn{4}{p{4 cm}}{\textit{Enhancement Layer}} & \\

 Skip Mask & 1.1M & 2.67\% & 0.02M/px & 1.33\%\\
 Ad. CANF & 13.6M & 34.27\% & 0.80M/px & 53.08\%\\

 \hline
 Total & 39.9M & & 1.50M/px &\\
  \hline
\end{tabular}
 \vspace{-0.15cm}
 \caption{A breakdown analysis of the model size and MACs for the encoder components (frame interpolator, base layer, enhancement layer).}
\vspace{-0.45cm}
 \label{tab:model-complexity-breakdown}
\end{table}

One critical element making our scheme successful is that we introduce a low-bitrate base layer that conveys the locations and  values of the unpredictable pixels. 
One significant advantage of the proposed scheme is its low computational complexity, particularly at the encoder. 
Compared to the state-of-the-art learned B-frame codec \cite{chen2022bcanf} with similar coding components, our scheme has an RD performance slightly lower at high bitrates and about the same at low bitrates. On the other hand, our approach uses only 55\% MACs operations in encoding and 73\% MACs in decoding. 
This is the first attempt at designing a two-layer video compression scheme
without motion coding. When the multi-frame merging network is replaced by a frame synthesis, the RD performance can be further improved as described in the supplementary document. Hence, there is a good potential to further improve its performance by tuning the parameters and altering the network architecture.  

\section{Acknowledgement}
This work is partially supported by MediaTek and the National Science and Technology Council, Taiwan (under Grant MOST 110-2221-E-A49 -065 and MOST 110-2634-F-A49-006). We would like to thank National Center for High-performance Computing (NCHC), Taiwan, for providing computational and storage resources for our experiments, and Mu-Jung Chen for his feedback and support. 


{\small
\bibliographystyle{ieee_fullname}
\bibliography{egbib}

\begin{thebibliography}{10}\itemsep=-1pt

\bibitem{FFMPEG}
Ffmpeg, 2022.
\newblock https://www.ffmpeg.org/.

\bibitem{fvcore}
Fvcore, 2022.
\newblock https://github.com/facebookresearch/fvcore.

\bibitem{HM1623}
Hm reference software for hevc, 2022.
\newblock https://vcgit.hhi.fraunhofer.de/jvet/HM/-/tree/HM-16.23/.

\bibitem{agustsson2020scale}
Eirikur Agustsson, David Minnen, Nick Johnston, Johannes Balle, Sung~Jin Hwang,
  and George Toderici.
\newblock Scale-space flow for end-to-end optimized video compression.
\newblock In {\em CVPR}, pages 8503--8512, 2020.

\bibitem{alexandre2022two}
David Alexandre, Hsueh-Ming Hang, and Wen-Hsiao Peng.
\newblock Two-layer learning-based p-frame coding with super-resolution and
  content-adaptive conditional anf.
\newblock In {\em Proceedings of the 4th ACM International Conference on
  Multimedia in Asia}, pages 1--7, 2022.

\bibitem{Bjontegaard2001}
Gisle Bjontegaard.
\newblock Calculation of average psnr differences between rd-curves.
\newblock {\em Document VCEG-M33}, 2001.

\bibitem{bossen2013common}
Frank Bossen.
\newblock Common test conditions and software reference configurations.
\newblock {\em JCTVC-L1100}, 12(7), 2013.

\bibitem{ccetin2022flexible}
Eren {\c{C}}etin, M~Ak{\i}n Y{\i}lmaz, and A~Murat Tekalp.
\newblock Flexible-rate learned hierarchical bi-directional video compression
  with motion refinement and frame-level bit allocation.
\newblock In {\em ICIP}, pages 1206--1210. IEEE, 2022.

\bibitem{chen2022learning}
Meixu Chen, Todd Goodall, Anjul Patney, and Alan~C Bovik.
\newblock Learning to compress videos without computing motion.
\newblock {\em Signal Processing: Image Communication}, 103:116633, 2022.

\bibitem{chen2022bcanf}
Mu-Jung Chen, Yi-Hsin Chen, Peng-Yu Chen, Chih-Hsuan Lin, Yung-Han Ho, and
  Wen-Hsiao Peng.
\newblock B-canf: Adaptive b-frame coding with conditional augmented
  normalizing flows.
\newblock {\em arXiv:2209.01769v1}, 2022.

\bibitem{cheng2019learning}
Zhengxue Cheng, Heming Sun, Masaru Takeuchi, and Jiro Katto.
\newblock Learning image and video compression through spatial-temporal energy
  compaction.
\newblock In {\em CVPR}, pages 10071--10080, 2019.

\bibitem{djelouah2019neural}
Abdelaziz Djelouah, Joaquim Campos, Simone Schaub-Meyer, and Christopher
  Schroers.
\newblock Neural inter-frame compression for video coding.
\newblock In {\em ICCV}, pages 6421--6429, 2019.

\bibitem{feng2020learned}
Runsen Feng, Yaojun Wu, Zongyu Guo, Zhizheng Zhang, and Zhibo Chen.
\newblock Learned video compression with feature-level residuals.
\newblock In {\em CVPR Workshop}, pages 120--121, 2020.

\bibitem{ho2021anfic}
Yung-Han Ho, Chih-Chun Chan, Wen-Hsiao Peng, Hsueh-Ming Hang, and Marek
  Doma{\'n}ski.
\newblock Anfic: Image compression using augmented normalizing flows.
\newblock {\em IEEE OJCAS}, 2:613--626, 2021.

\bibitem{ho2022canf}
Yung-Han Ho, Chih-Peng Chang, Peng-Yu Chen, Alessandro Gnutti, and Wen-Hsiao
  Peng.
\newblock Canf-vc: Conditional augmented normalizing flows for video
  compression.
\newblock In {\em ECCV}, pages 207--223. Springer, 2022.

\bibitem{hu2020improving}
Zhihao Hu, Zhenghao Chen, Dong Xu, Guo Lu, Wanli Ouyang, and Shuhang Gu.
\newblock Improving deep video compression by resolution-adaptive flow coding.
\newblock In {\em ECCV}, pages 193--209. Springer, 2020.

\bibitem{hu2022coarse}
Zhihao Hu, Guo Lu, Jinyang Guo, Shan Liu, Wei Jiang, and Dong Xu.
\newblock Coarse-to-fine deep video coding with hyperprior-guided mode
  prediction.
\newblock In {\em CVPR}, pages 5921--5930, 2022.

\bibitem{hu2021fvc}
Zhihao Hu, Guo Lu, and Dong Xu.
\newblock Fvc: A new framework towards deep video compression in feature space.
\newblock In {\em CVPR}, pages 1502--1511, 2021.

\bibitem{huang2022real}
Zhewei Huang, Tianyuan Zhang, Wen Heng, Boxin Shi, and Shuchang Zhou.
\newblock Real-time intermediate flow estimation for video frame interpolation.
\newblock In {\em ECCV}, pages 624--642. Springer, 2022.

\bibitem{kingma2014adam}
Diederik~P Kingma and Jimmy Ba.
\newblock Adam: A method for stochastic optimization.
\newblock 2015.

\bibitem{ladune2021conditional}
Th{\'e}o Ladune, Pierrick Philippe, Wassim Hamidouche, Lu Zhang, and Olivier
  D{\'e}forges.
\newblock Conditional coding for flexible learned video compression.
\newblock {\em ICLR}, 2021.

\bibitem{li2021deep}
Jiahao Li, Bin Li, and Yan Lu.
\newblock Deep contextual video compression.
\newblock {\em NeurIPS}, 34, 2021.

\bibitem{li2022hybrid}
Jiahao Li, Bin Li, and Yan Lu.
\newblock Hybrid spatial-temporal entropy modelling for neural video
  compression.
\newblock In {\em ACM Multimedia}, 2022.

\bibitem{lin2020m}
Jianping Lin, Dong Liu, Houqiang Li, and Feng Wu.
\newblock M-lvc: Multiple frames prediction for learned video compression.
\newblock In {\em CVPR}, pages 3546--3554, 2020.

\bibitem{lu2020content}
Guo Lu, Chunlei Cai, Xiaoyun Zhang, Li Chen, Wanli Ouyang, Dong Xu, and Zhiyong
  Gao.
\newblock Content adaptive and error propagation aware deep video compression.
\newblock In {\em ECCV}, pages 456--472. Springer, 2020.

\bibitem{lu2019dvc}
Guo Lu, Wanli Ouyang, Dong Xu, Xiaoyun Zhang, Chunlei Cai, and Zhiyong Gao.
\newblock Dvc: An end-to-end deep video compression framework.
\newblock In {\em CVPR}, pages 11006--11015, 2019.

\bibitem{lu2020end}
Guo Lu, Xiaoyun Zhang, Wanli Ouyang, Li Chen, Zhiyong Gao, and Dong Xu.
\newblock An end-to-end learning framework for video compression.
\newblock {\em IEEE TPAMI}, 43(10):3292--3308, 2020.

\bibitem{mercat2020uvg}
Alexandre Mercat, Marko Viitanen, and Jarno Vanne.
\newblock Uvg dataset: 50/120fps 4k sequences for video codec analysis and
  development.
\newblock In {\em MMSys}, pages 297--302, 2020.

\bibitem{pourreza2021extending}
Reza Pourreza and Taco Cohen.
\newblock Extending neural p-frame codecs for b-frame coding.
\newblock In {\em ICCV}, pages 6680--6689, 2021.

\bibitem{rocca1969}
Fabio Rocca.
\newblock Television bandwidth compression utilizing frame-to-frame correlation
  and movement compensation.
\newblock In {\em Symposium on Picture Bandwidth Compression}. Massachusetts
  Institute of Technology, 1969.

\bibitem{Sullivan2012hevc}
Gary~J Sullivan, Jens-Rainer Ohm, Woo-Jin Han, and Thomas Wiegand.
\newblock Overview of the high efficiency video coding (hevc) standard.
\newblock {\em IEEE TCSVT}, 22(12):1649--1668, 2012.

\bibitem{wu2018video}
Chao-Yuan Wu, Nayan Singhal, and Philipp Krahenbuhl.
\newblock Video compression through image interpolation.
\newblock In {\em ECCV}, pages 416--431, 2018.

\bibitem{xue2019video}
Tianfan Xue, Baian Chen, Jiajun Wu, Donglai Wei, and William~T Freeman.
\newblock Video enhancement with task-oriented flow.
\newblock {\em IJCV}, 127(8):1106--1125, 2019.

\bibitem{yang2020Learning}
Ren Yang, Fabian Mentzer, Luc Van~Gool, and Radu Timofte.
\newblock Learning for video compression with hierarchical quality and
  recurrent enhancement.
\newblock In {\em CVPR}, 2020.

\bibitem{yilmaz2021end}
M~Ak{\i}n Y{\i}lmaz and A~Murat Tekalp.
\newblock End-to-end rate-distortion optimized learned hierarchical
  bi-directional video compression.
\newblock {\em IEEE TIP}, 31:974--983, 2021.

\bibitem{zou2021adaptation}
Nannan Zou, Honglei Zhang, Francesco Cricri, Hamed~R Tavakoli, Jani Lainema,
  Emre Aksu, Miska Hannuksela, and Esa Rahtu.
\newblock Adaptation and attention for neural video coding.
\newblock In {\em ISM}, pages 240--244, 2021.

\bibitem{zou2021learned}
Nannan Zou, Honglei Zhang, Francesco Cricri, Hamed~R Tavakoli, Jani Lainema,
  Emre Aksu, Miska Hannuksela, and Esa Rahtu.
\newblock Learned video compression with intra-guided enhancement and implicit
  motion information.
\newblock In {\em ICCV}, pages 1870--1874, 2021.

\end{thebibliography}
}

\end{document}